\documentclass{article} 
\usepackage[final]{colm2026_conference}

\usepackage{microtype}
\usepackage{hyperref}
\usepackage{url}
\usepackage{booktabs}
\usepackage{graphicx}
\usepackage{algorithm}
\usepackage{algpseudocode}
\usepackage{amsmath}
\usepackage{amssymb}
\usepackage{multirow}
\usepackage{subcaption}
\usepackage{wrapfig}
\usepackage[table]{xcolor}
\usepackage{enumitem}

\definecolor{darkblue}{rgb}{0, 0, 0.5}
\hypersetup{colorlinks=true, citecolor=darkblue, linkcolor=darkblue, urlcolor=darkblue}

\title{Beyond Trajectory Imitation: Strategy-Guided Policy Optimization for LLM Reasoning}

\author{Tianyuan Shi\textsuperscript{1},
Canbin Huang\textsuperscript{1},
Bei Li\textsuperscript{2},
Xin Chen\textsuperscript{2},
Xiaojun Quan\textsuperscript{3}\thanks{Corresponding author},
Jingang Wang\textsuperscript{2}
\\
\textbf{Qifan Wang}\textsuperscript{4}
\\
\textsuperscript{1}School of Computer Science and Engineering, Sun Yat-sen University, China
\\
\textsuperscript{2}Meituan, Inc., China,
\textsuperscript{3}Shenzhen Loop Area Institute, China,
\textsuperscript{4}Meta AI, USA
\\
\texttt{\{shity6,huangcb3\}@mail2.sysu.edu.cn},
\texttt{xiaojunquan@slai.edu.cn}
\\
\texttt{\{libei17,chenxin148,wangjingang02\}@meituan.com},
\texttt{wqfcr@fb.com}
}

\begin{document}

\ifcolmsubmission
\fi

\maketitle

\begin{abstract}
Distilling reasoning capabilities from strong to weak language models
typically involves imitating specific solution trajectories, effectively
transferring \emph{what to answer} rather than \emph{how to reason}.
This trajectory-level imitation encourages memorization of
instance-specific steps rather than acquisition of transferable
problem-solving skills, limiting generalization to novel problems.
We propose \textbf{Strategy-Guided Policy Optimization} (SGPO), which
replaces instance-level trajectory imitation with reusable
\emph{strategy distillation}.
SGPO extracts structured strategy descriptions from strong-model
responses and, for each problem, constructs both autonomous and
strategy-guided trajectories to enable direct comparison of the model's
behavior with and without strategic guidance. The framework then
addresses two key questions. For \emph{how} to distill, a token-level
forward-KL objective selectively transfers the distributional shift
induced by strategy conditioning into the unguided policy, with proximal
constraints ensuring stability. For \emph{when} to distill, adaptive
instance-level weighting strengthens guidance when autonomous
exploration falls short and reduces it as the model's own competence
grows. Experiments on four mathematical benchmarks across two model
families show that SGPO consistently outperforms SFT, on-policy RL,
and hybrid-policy baselines, improving the average
score by 2.2 points over the strongest baseline on Qwen2.5-7B-Instruct.
Analysis reveals that the forward-KL objective provides an inherently
selective distillation signal that outperforms direct trajectory
imitation, and that strategy distillation exhibits complementary scaling
with base model capability. Code is available at
\url{https://github.com/Stycoo/SGPO}.
\end{abstract}

\section{Introduction}
\label{sec:intro}

Large language models (LLMs) have demonstrated remarkable reasoning
capabilities~\citep{deepseekr1,openai_o1}, motivating a range of
methods to transfer these capabilities to smaller models. Whether
through supervised fine-tuning on expert
traces~\citep{deepseekr1,limo,generalizationsft,psft} or hybrid
objectives that integrate expert demonstrations into policy
optimization~\citep{luffy,hpt,srft}, existing approaches operate on the
same unit of knowledge transfer: \emph{instance-level solution
trajectories}. The student is trained to reproduce what the expert
wrote, that is, specific sequences of reasoning steps for specific
problems, but is never taught the reusable problem-solving strategy that
explains \emph{why} those steps were chosen. This trajectory-level
imitation encourages memorization of instance-specific patterns rather
than acquisition of transferable skills, limiting generalization to
novel problems. A natural question arises: \emph{Can we shift the
distillation target from specific solutions to the reusable
problem-solving strategies that make those solutions effective?}

We propose \textbf{Strategy-Guided Policy Optimization} (SGPO), a
framework designed to accomplish this shift. SGPO extracts structured
\emph{strategy descriptions} from strong-model responses. Each
description specifies the problem type, the solving approach, and the
general procedural steps, without carrying out computations or revealing
the answer. Rather than asking the student to reproduce what the expert
generates, SGPO uses these descriptions to reshape the student's own
reasoning distribution, distilling \emph{how to reason} rather than
\emph{what to answer}.

A central design principle is that strategic knowledge should be
\emph{internalized} into the model's unguided policy rather than remain
an external dependency unavailable at inference time. To this end,
SGPO constructs both autonomous and strategy-guided trajectories for
each problem~(\S\ref{sec:formulation}), enabling direct comparison of
the model's behavior with and without strategic guidance. This dual
construction serves as the foundation for addressing two complementary
questions: 
(1) \textbf{How to distill.} A token-level forward-KL objective measures the divergence between guided and unguided next-token distributions and selectively distills strategy-critical information back into the unguided policy, while proximal constraints at both trajectory and token levels ensure stability~(\S\ref{sec:kl_distill}, \S\ref{sec:proximal}). (2) \textbf{When to distill.} Adaptive per-instance weighting modulates distillation strength based on
the marginal benefit of strategy guidance, strengthening distillation
when autonomous exploration falls short and reducing it as the model's
own competence grows~(\S\ref{sec:proximal}). This naturally transitions
training from strategy-driven rapid improvement in early stages to
autonomous-policy-dominated steady optimization in later stages, without
manual scheduling.



Crucially, SGPO never imitates any trajectory, whether autonomous or
guided. Instead, it distills the \emph{distributional shift} caused by
strategy conditioning, operating at the level of token-level probability
changes rather than sequence matching. This enables selective transfer
of strategic knowledge while preserving the reasoning diversity the
model has already acquired through autonomous exploration.

Experiments on four mathematical benchmarks across two model families
(Qwen2.5 and Llama-3.1) show that SGPO consistently outperforms
strong baselines including SFT, on-policy RL, and state-of-the-art
hybrid-policy methods, improving the average score by 2.2 points over
the strongest baseline on Qwen2.5-7B-Instruct. Analysis reveals two
findings. First, the forward-KL objective provides an inherently
selective distillation signal: without any token-level annotation,
optimization pressure concentrates on tokens whose probability shifts
most under strategy conditioning, which empirically correspond to
strategy-critical decision points rather than routine linguistic tokens.
Second, strategy distillation exhibits complementary scaling with base
model capability: as foundational reasoning competence grows, the
ability to internalize strategic guidance increases at a faster rate,
suggesting that a minimum reasoning competence is needed to benefit from
strategy-level transfer.

\section{Related Work}
\subsection{Supervised Fine-Tuning for LLM Reasoning}

Training a weaker model on expert reasoning traces is the most common
form of reasoning distillation~\citep{deepseekr1,limo}.  While simple
and effective, this trajectory-level distillation is sensitive to data
quality~\citep{limo}, prone to memorization~\citep{sftmemorizes}, and
vulnerable to exposure bias~\citep{generalizationsft}.  Recent work
mitigates these issues from RL perspectives along two directions.
The first introduces proximal constraints:
\citet{generalizationsft} downweight the loss on well-learned tokens
using the model's own predictions, while
PSFT~\citep{psft} applies proximal clipping to bound update magnitudes.
The second recasts SFT through a reward-optimization lens; for example,
IW-SFT~\citep{iw-sft} tightens the RL lower bound via importance
weighting on policy probability ratios.

These methods improve \emph{how} the teacher's output is transferred but
do not change \emph{what} is transferred: the student still imitates
specific solutions.  In contrast, our work operates at a higher level of
abstraction, distilling reusable \emph{strategies} rather
than concrete trajectories.

\subsection{Hybrid Policy Optimization for LLM Reasoning}

Incorporating expert demonstrations into policy optimization is a
long-standing theme in RL~\citep{dapg, nair2018overcoming}.  Recent LLM
methods follow two directions.
\textbf{Unified-loss methods} mix expert data into the RL objective:
LUFFY~\citep{luffy} adds off-policy expert trajectories to GRPO rollout
groups with regularized importance sampling;
AMPO~\citep{ampo} replaces incorrect on-policy samples with diverse
off-policy alternatives;
other approaches interleave RL and SFT
updates~\citep{ReLIFT,chen2025schedule};
and SRFT~\citep{srft} unifies these ideas in a single-stage framework
with sample-level modulation.
\textbf{Prefix-guided methods} use expert trajectories to structure
generation: UFT~\citep{uft} progressively masks expert suffixes to
encourage autonomy, while BREAD~\citep{bread} branches rollouts from
intermediate expert steps.

All these methods transfer knowledge at the level of specific solution
steps.  Our approach differs in two respects:
(i)~it conditions on strategy descriptions that specify problem-solving
direction without prescribing concrete computations, and
(ii)~rather than imitating the teacher's output, it distills the
\emph{distributional shift} that strategy conditioning induces in the
student's own policy, making the transfer inherently adapted to the
student's current capability while removing dependence on external hints
at inference time.

\section{Method}
\label{sec:method}

\subsection{Problem Formulation and Overview}
\label{sec:formulation}

Let $\pi_{\theta}(\cdot \mid q)$ denote the policy of the target model
for a reasoning problem $q$.  For each problem, we assume access to a
\emph{strategy description} $s$ extracted from a strong-model response.
A strategy description is a concise natural-language summary that
specifies the problem type, the solving approach, and the general
procedural steps, without carrying out intermediate computations or
revealing the final answer.  It encodes actionable strategic information
while omitting solution-specific details, occupying a middle ground
between a generic hint and a complete solution.  Details of the
extraction pipeline and prompt templates are
provided in Appendix~\ref{app:strategy_extraction}.

For each training problem $q$, we construct two trajectory groups:
\textbf{(1)~Autonomous group} $\{o_i\}_{i=1}^{G_1}$, sampled
from $\pi_{\theta}(\cdot \mid q)$; and
\textbf{(2)~Strategy-guided group}
$\{\tilde{o}_j\}_{j=1}^{G_2}$, sampled from
$\pi_{\theta}(\cdot \mid q, s)$ where $s$ is prepended to the prompt.
The central challenge is to convert the information gained under the
guided condition into reusable reasoning ability under the original
unguided condition.  The remainder of this section describes three
mechanisms that jointly address this challenge: autonomous GRPO
optimization (\S\ref{sec:grpo}), token-level forward-KL distillation
(\S\ref{sec:kl_distill}), and proximal constraints with adaptive
weighting (\S\ref{sec:proximal}).

\begin{figure*}[t]
    \centering
    \includegraphics[width=\textwidth]{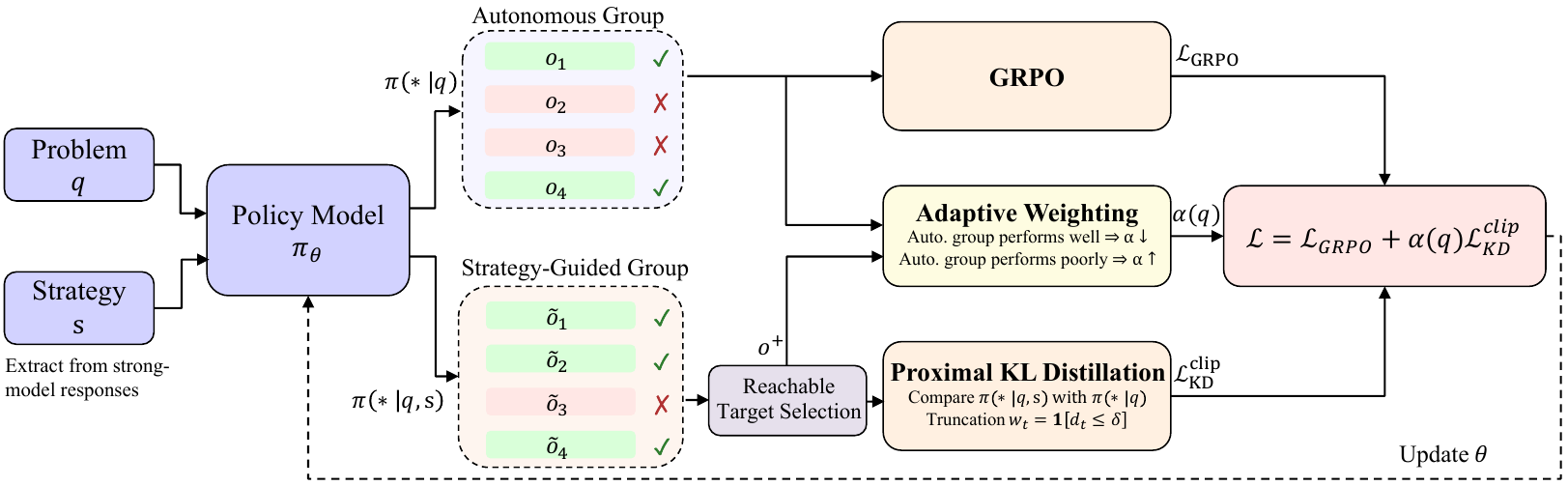}
    \caption{Overview of the SGPO framework.  For each problem, we
    jointly construct an autonomous group and a strategy-guided group.
    The autonomous group is optimized with GRPO\@.  Correct trajectories
    from the strategy-guided group provide proximal KL distillation
    signals to the unguided policy.  An adaptive weight $\alpha(q)$
    controls distillation strength.}
    \label{fig:framework}
\end{figure*}

\subsection{Autonomous Exploration with GRPO}
\label{sec:grpo}

The autonomous trajectory group is optimized with Group Relative Policy
Optimization (GRPO; \citealt{grpo}).  For each problem $q$, the $G_1$
sampled responses are scored by a verifiable reward function
$R(o_i, q) \in \{0,1\}$.  Rewards within each group are normalized to
zero-mean, unit-variance advantages $\hat{A}_i$, and the policy is
updated via the clipped objective:
\begin{equation}
\mathcal{L}_{\mathrm{GRPO}}(\theta)
= \mathbb{E}_{q,\,\{o_i\}}
\!\Bigg[
\frac{1}{G_1}\sum_{i=1}^{G_1}\frac{1}{|o_i|}
\sum_{t=1}^{|o_i|}
\min\!\Big(
  \rho_{i,t}\,\hat{A}_{i},\;
  \mathrm{clip}\big(\rho_{i,t},1{-}\varepsilon,1{+}\varepsilon\big)
  \hat{A}_{i}
\Big)
\Bigg],
\label{eq:grpo}
\end{equation}
where
$\rho_{i,t}
={\pi_{\theta}(o_{i,t}\mid q,o_{i,<t})}\big/
 {\pi_{\theta_{\mathrm{old}}}(o_{i,t}\mid q,o_{i,<t})}$
is the token-level importance ratio.  This objective ensures continued
improvement on the model's own distribution independently of strategy
guidance.

\subsection{Strategy-Guided Distillation}
\label{sec:kl_distill}

Rather than directly applying SFT to successful guided trajectories—
which would collapse back into trajectory imitation and fail to separate
strategic knowledge from surface realization—we construct a distillation
signal that captures how strategy conditioning shifts the model's own
token-level distribution.

Concretely, let
$\mathcal{C}=\{j : R(\tilde{o}_j,q)=1\}$ be the set of correct guided
trajectories.  We define the token-level KL distillation objective over
$\mathcal{C}$ as:
\begin{equation}
\mathcal{L}_{\mathrm{KD}}(\theta)
=
\frac{1}{|\mathcal{C}|}
\sum_{j \in \mathcal{C}}
\frac{1}{|\tilde{o}_j|}
\sum_{t=1}^{|\tilde{o}_j|}
d_{j,t}(\theta),
\label{eq:kd_basic}
\end{equation}
where
\begin{equation}
d_{j,t}(\theta)
=
D_{\mathrm{KL}}
\!\left(
\operatorname{sg}\!\left[\pi_\theta(\cdot\mid q,s,\tilde{o}_{j,<t})\right]
\;\middle\|\;
\pi_\theta(\cdot\mid q,\tilde{o}_{j,<t})
\right).
\label{eq:token_kl}
\end{equation}

\begin{wrapfigure}{r}{0.52\textwidth}
    \centering
    \includegraphics[width=0.52\textwidth]{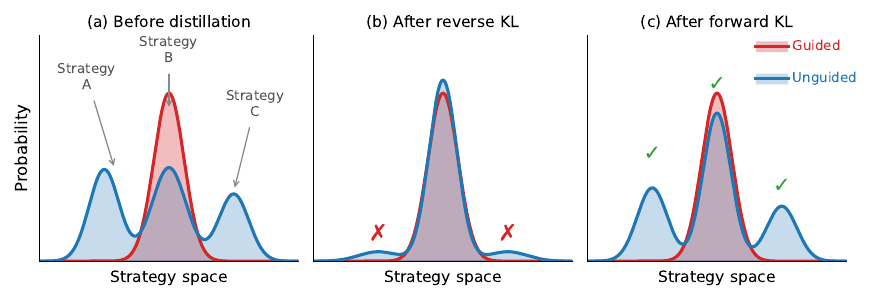}
    \caption{KL direction.  (a)~The unguided policy covers multiple
    strategies; the guided distribution reflects one.  (b)~Reverse KL
    collapses onto the guided mode.  (c)~Forward KL absorbs the guided
    strategy while preserving alternatives.}
    \label{fig:kl_direction}
    \vspace{-10pt}
\end{wrapfigure}

\emph{First}, we adopt the forward KL direction (guided as reference).
The strategy-guided distribution concentrates on a particular strategy,
whereas the unguided policy may cover multiple effective strategies
discovered through autonomous exploration
(Figure~\ref{fig:kl_direction}).  Reverse KL would actively penalize
mass outside the guided mode, collapsing the policy onto a single
strategy.  Forward KL only encourages coverage of the guided mode
without suppressing alternatives, and together with the GRPO objective
that continuously reinforces independently discovered strategies,
preserves existing reasoning diversity while absorbing new strategic
insight (empirical comparison in
Appendix~\ref{app:kl_direction_ablation}).
\emph{Second}, the stop-gradient $\operatorname{sg}[\cdot]$ restricts
gradients to the unguided distribution.  Both distributions in
Eq.~\eqref{eq:token_kl} share the same output prefix
$\tilde{o}_{j,<t}$ but differ in prompt context: $(q, s)$ for the
guided side vs.\ $q$ alone for the unguided side.  Without the
stop-gradient, the shared $\theta$ could trivially reduce the KL by
shifting the guided distribution toward the unguided one.

The resulting objective has a natural selectivity property: at each token
position, the KL value is large when the next-token distribution shifts
substantially under strategy conditioning (e.g., choosing to apply the
quadratic formula vs.\ factoring), and small when the token mainly
reflects generic linguistic realization (e.g., formatting).~(\S\ref{sec:kl_distill})
Distillation pressure thus automatically concentrates on positions that
encode strategic decisions, without requiring any explicit token-level
annotation.

\subsection{Proximal Constraints and Adaptive Weighting}
\label{sec:proximal}

Directly minimizing the KL term in Eq.~\eqref{eq:kd_basic} can be
unstable when the guided and unguided distributions diverge
substantially, risking entropy collapse (\S\ref{sec:ablation}).  We
introduce proximal constraints for training stability and an adaptive
weighting mechanism for distillation efficiency, operating at three
complementary levels.

\textbf{Trajectory level: reachable target selection.}\quad
Rather than averaging over all correct guided trajectories as in
Eq.~\eqref{eq:kd_basic}, we select the single trajectory most reachable
by the current unguided policy:
\begin{equation}
o^{+}
= \arg\max_{\tilde{o}_j \in \mathcal{C}}
\sum_{t=1}^{|\tilde{o}_j|}
\log \pi_\theta(\tilde{o}_{j,t} \mid q, \tilde{o}_{j,<t}).
\label{eq:reachable_select}
\end{equation}
Here we evaluate the guided trajectory under the \emph{unguided} policy
(without $s$ in the conditioning) to measure how easily the model could
produce this trajectory on its own. This preserves correctness while
minimizing the distributional gap that the distillation step must
bridge. When $|\mathcal{C}|=0$, the distillation term is skipped
entirely.

\textbf{Token level: KL clipping.}\quad
We mask token positions where the guided--unguided divergence exceeds a
threshold $\delta$, yielding the clipped objective over the selected
trajectory:
\begin{equation}
\mathcal{L}_{\mathrm{KD}}^{\mathrm{clip}}(\theta)
=
\frac{1}{|o^+|}
\sum_{t=1}^{|o^+|}
w_{t}\, d_{t}(\theta),
\quad
w_{t}=\mathbf{1}\!\left[d_{t}(\theta)\le \delta\right],
\label{eq:kd_clip}
\end{equation}
where $d_t(\theta)$ denotes the per-token KL from
Eq.~\eqref{eq:token_kl} evaluated on $o^+$.  This prevents outlier
positions from dominating optimization.

\textbf{Instance level: adaptive weighting.}\quad
Let $p_{\text{auto}}=\frac{1}{G_1}\sum_i R(o_i,q)$ and
$p_{\text{guide}}=\frac{1}{G_2}\sum_j R(\tilde{o}_j,q)$ denote the
autonomous and guided pass rates.  We modulate the per-instance
distillation strength as:
\begin{equation}
\alpha(q) = \mathrm{clip}\!\left(
\alpha_0 \cdot
\frac{p_{\text{guide}} - p_{\text{auto}} + \gamma}
     {p_{\text{auto}} + \gamma},
\; 0, \; \alpha_{\max}\right),
\label{eq:adaptive_weight}
\end{equation}
where $\alpha_0$ is a base coefficient, $\gamma>0$ a smoothing constant,
and $\alpha_{\max}$ a cap.  The weight is large when strategy guidance
substantially raises the pass rate and vanishes when the model already
solves the problem autonomously. 
A comparison with alternative weighting strategies is given in
Appendix~\ref{app:weight_ablation}.

\subsection{Overall Objective}
\label{sec:overall}

For each problem $q$, the training loss combines autonomous GRPO with
proximal strategy distillation:
\begin{equation}
\mathcal{L}(\theta; q)
=
\mathcal{L}_{\mathrm{GRPO}}(\theta; q)
\;+\;
\alpha(q)\,\mathcal{L}_{\mathrm{KD}}^{\mathrm{clip}}(\theta; q),
\label{eq:total_loss}
\end{equation}
and is averaged over the mini-batch.  Conceptually, strategy extraction
determines \emph{what} to transfer; the token-level forward-KL
objective identifies \emph{where} to transfer by concentrating on
strategy-critical positions; and the three-level proximal constraints
control \emph{how much} to transfer at the trajectory, token, and
instance granularity.  The full training algorithm is given in
Appendix~\ref{app:algorithm}.

\section{Experimental Setup}

\subsection{Models and Data}

We conduct experiments on two model families:
Qwen2.5-\{1.5B, 7B\}-Instruct~\citep{qwen2.5} and
Llama-3.1-8B-Instruct~\citep{llama3}, covering both different
architectures and different scales. Training data consist of 8.5K
problems randomly sampled from the LUFFY dataset~\citep{luffy}, with
reference solutions generated by DeepSeek-R1. Each RL training instance
requires $(G_1{+}G_2){=}12$ sampled trajectories, yet stable gains are
already observed at this moderate scale, indicating favorable data
efficiency. Strategy descriptions are extracted from the
corresponding reference solutions via the procedure described in
Appendix~\ref{app:strategy_extraction}. An analysis of strategy description quality is provided in Appendix~\ref{app:method_quality}.

\subsection{Evaluation Benchmarks}

We evaluate on four mathematical reasoning benchmarks of increasing difficulty: MATH500~\citep{math500}, AMC~23, OlympiadBench~\citep{olympiadbench} and AIME~24. Because AMC~23 and AIME~24 are relatively small test sets, we report avg@32 (the accuracy averaged over 32 independent samples per problem) to reduce variance; for the larger MATH500 and OlympiadBench, we report avg@4. All evaluations use a maximum generation length of 32K tokens, temperature 0.6, and top-$p{=}0.95$.

\subsection{Baselines}

We compare against four baselines. \textbf{SFT} performs supervised
fine-tuning on expert reasoning traces only. \textbf{SFT\,+\,GRPO}
follows SFT with standard on-policy GRPO. These two baselines test
whether strategy distillation provides gains beyond conventional
supervised and reinforcement learning. \textbf{HPT}~\citep{hpt}
combines on-policy RL with an auxiliary SFT loss on expert trajectories
within a unified objective. \textbf{LUFFY}~\citep{luffy} incorporates
off-policy expert trajectories into GRPO rollout groups with regularized
importance sampling. HPT and LUFFY are the most directly relevant
comparisons, as they also leverage strong-model knowledge during policy
optimization. 

\subsection{Implementation Details}

We split the 8.5K training problems into roughly one third for SFT
warm-up and the remainder for RL training. The SFT stage uses learning
rate $1{\times}10^{-5}$ for 10 epochs to equip base models with basic
long-chain reasoning ability. The RL stage uses learning rate
$1{\times}10^{-6}$ for 10 epochs with total group size 12 ($G_1{=}8$
autonomous, $G_2{=}4$ strategy-guided). Key hyperparameters include GRPO
clipping bounds $\varepsilon_{\text{low}}{=}0.2$ and
$\varepsilon_{\text{high}}{=}0.28$~\citep{dapo}, KL threshold $\delta{=}1.0$, base
adaptive coefficient $\alpha_0{=}0.5$, smoothing constant
$\gamma{=}0.1$, and maximum adaptive weight $\alpha_{\max}{=}0.8$. All
baselines share the same SFT warm-up configuration and total sample
budget to ensure fair comparison. Full details are provided in
Appendix~\ref{app:impl_details}.

\section{Results}

\subsection{Main Results}

We summarize the main results in Table~\ref{tab:reasoning_main_results} and discuss two key observations.

\textbf{Consistent improvements over all baselines.}
SGPO achieves the best average performance across all three base models
and both model families. Notably, HPT and LUFFY also leverage
strong-model knowledge during RL, yet SGPO consistently outperforms
them by transferring knowledge at the strategy level rather than the
trajectory level. The mechanism underlying this advantage is examined
in \S\ref{sec:selective}.

\textbf{Complementary scaling with base model capability.}
The gains from SGPO scale with the strength of the base model: on
Qwen2.5-7B-Instruct, the average score improves by 9.9 points over
the base model (42.2$\to$52.1), compared with 6.7 points for
Qwen2.5-1.5B-Instruct (23.2$\to$29.9). This suggests that stronger
models are better able to convert strategy descriptions into effective
guided exploration and internalize the resulting gains into the
unguided policy.

\begin{table*}[htbp]
\centering
\caption{Main results on four mathematical reasoning benchmarks.
Average is the arithmetic mean of the four benchmark scores.}
\label{tab:reasoning_main_results}
\small
\setlength{\tabcolsep}{6pt}
\resizebox{0.85\textwidth}{!}{
\begin{tabular}{lccccc}
\toprule
\textbf{Model} & \textbf{MATH500} & \textbf{AMC23} & \textbf{Olympiad} & \textbf{AIME24} & \textbf{Average} \\
\midrule
Qwen2.5-1.5B-Instruct & 50.1 & 21.3 & 18.4 & 3.0 & 23.2 \\
\quad SFT & 48.4 & 19.7 & 16.1 & 4.7 & 22.2 \\
\quad SFT+GRPO & 54.1 & 26.2 & 21.7 & 5.5 & 26.9 \\
\quad HPT & 53.8 & 28.5 & 22.6 & 7.7 & 28.2 \\
\quad LUFFY & 54.4 & 27.9 & 23.1 & 8.4 & 28.5 \\
\rowcolor{blue!8} \quad SGPO & \textbf{57.7} & \textbf{29.0} & \textbf{23.7} & \textbf{9.0} & \textbf{29.9} \\
\midrule
Qwen2.5-7B-Instruct & 75.2 & 43.0 & 38.8 & 11.8 & 42.2 \\
\quad SFT & 76.4 & 42.1 & 40.0 & 12.4 & 42.7 \\
\quad SFT+GRPO & 80.3 & 52.7 & 48.4 & 18.0 & 49.9 \\
\quad HPT & 79.1 & 53.5 & 48.8 & 17.0 & 49.6 \\
\quad LUFFY & 78.0 & 53.0 & 47.6 & 16.4 & 48.8 \\
\rowcolor{blue!8} \quad SGPO & \textbf{82.7} & \textbf{55.9} & \textbf{50.0} & \textbf{19.7} & \textbf{52.1} \\
\midrule
Llama-3.1-8B-Instruct & 43.7 & 20.3 & 14.5 & 3.0 & 20.4 \\
\quad SFT & 45.1 & 21.7 & 16.7 & 6.0 & 22.4 \\
\quad SFT+GRPO & 50.0 & 23.4 & 19.3 & 8.0 & 25.2 \\
\quad HPT & 51.3 & 24.0 & 20.1 & 9.8 & 26.3 \\
\quad LUFFY & 50.7 & 24.3 & 21.9 & 9.4 & 26.5 \\
\rowcolor{blue!8} \quad SGPO & \textbf{52.0} & \textbf{25.6} & \textbf{22.9} & \textbf{10.0} & \textbf{27.6} \\
\bottomrule
\end{tabular}
}
\end{table*}

\begin{table}[t]
\centering
\caption{Ablation study on Qwen2.5-7B-Instruct.}
\label{tab:ablation}
\small
\setlength{\tabcolsep}{6pt}
\resizebox{0.85\linewidth}{!}{
\begin{tabular}{lccccc}
\toprule
\textbf{Setting} & \textbf{MATH500} & \textbf{AMC23} & \textbf{Olympiad} & \textbf{AIME24} & \textbf{Average} \\
\midrule
\rowcolor{blue!8} SGPO & \textbf{82.7} & \textbf{55.9} & \textbf{50.0} & \textbf{19.7} & \textbf{52.1} \\
\midrule
w/o strategy distillation & 80.3 & 52.7 & 48.4 & 18.0 & 49.9 \\
w/o autonomous GRPO & 78.0 & 53.1 & 47.1 & 16.3 & 48.6 \\
\midrule
w/o all proximal constraints & 76.9 & 47.3 & 42.4 & 14.4 & 45.3 \\
w/o KL clipping & 79.2 & 52.1 & 47.1 & 16.3 & 48.7 \\
w/o target selection & 78.8 & 50.4 & 47.0 & 16.0 & 48.1 \\
\midrule
w/o adaptive weighting & 81.0 & 54.8 & 48.9 & 18.7 & 50.9 \\
\bottomrule
\end{tabular}
}
\vspace{-10pt}
\end{table}

\subsection{Ablation Studies}
\label{sec:ablation}
\begin{figure*}[htbp]
    \centering
    \includegraphics[width=\textwidth]{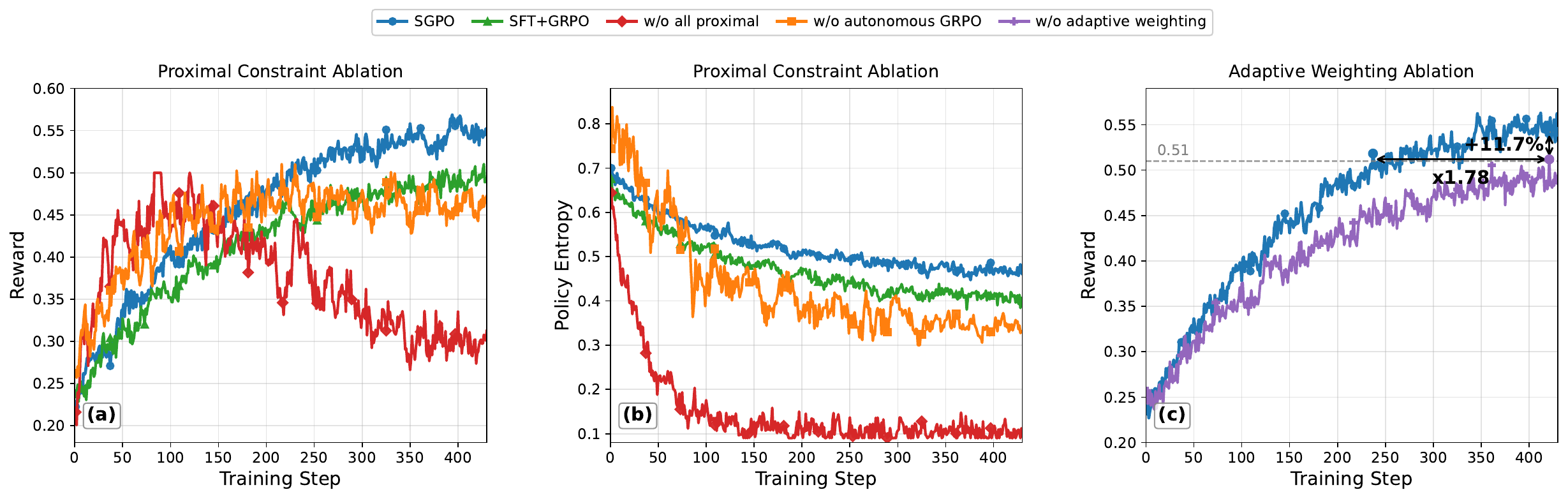}
    \caption{Training dynamics under ablation settings.
    (a)(b)~Removing either autonomous GRPO or all proximal constraints leads to a rapid entropy collapse.
    (c)~Adaptive weighting accelerates convergence and raises the final reward ceiling.
    }
    \label{fig:ablation_dynamics}
    \vspace{-5pt}
\end{figure*}

\vspace{-3pt}
\textbf{Autonomous GRPO and strategy distillation are
complementary.}\quad
Removing the strategy distillation term ($\alpha(q){=}0$, reducing
to SFT+GRPO) lowers the average score from 52.1 to 49.9, confirming
that strategy-level transfer provides gains beyond on-policy RL
alone. Removing autonomous GRPO (i.e., training with only the KL
distillation objective on strategy-guided trajectories) leads to a larger
drop (52.1$\to$48.6). As shown in
Figure~\ref{fig:ablation_dynamics}(a)(b), this setting yields fast
early reward growth but the progress is unsustainable: entropy drops
sharply and reward subsequently stagnates. A plausible explanation is
that without autonomous rollouts, the policy loses the distributional
diversity needed to explore beyond the strategies prescribed by the
extracted, causing optimization to plateau once the
easily exploitable strategy descriptions are exhausted. These results
demonstrate the necessity of combining both objectives: GRPO maintains
broad exploration while distillation injects targeted strategic
knowledge.

\textbf{Proximal constraints are essential for stable distillation.}\quad
Removing all proximal constraints causes the largest single-component
drop (52.1$\to$45.3, $-$6.8). Both sub-components contribute, with
reachable target selection ($-$4.0) slightly more impactful than KL
clipping ($-$3.4), indicating that whether the distillation target is
reachable has a particularly direct effect on performance. Unlike the
w/o GRPO setting where entropy declines gradually due to lack of
exploration diversity, removing proximal constraints triggers abrupt
entropy collapse driven by excessively large KL updates, as shown in
Figure~\ref{fig:ablation_dynamics}(a)(b). The two constraints address
complementary failure modes: target selection limits excessive
distributional shifts at the trajectory level, while KL clipping
suppresses local token-level outliers.

\textbf{Adaptive weighting improves efficiency and ceiling.}\quad
Removing adaptive weighting leads to a moderate drop
(52.1$\to$50.9). Figure~\ref{fig:ablation_dynamics}(c) further shows
that adaptive weighting accelerates optimization, reaching the same
reward level in fewer steps. This confirms that strategy guidance is
not equally useful across instances, and concentrating distillation
budget on high-marginal-benefit problems improves both convergence speed
and final performance. A comparison with alternative weighting
strategies is provided in Appendix~\ref{app:weight_ablation}.

\begin{figure*}[h]
    \centering
    \includegraphics[width=\textwidth]{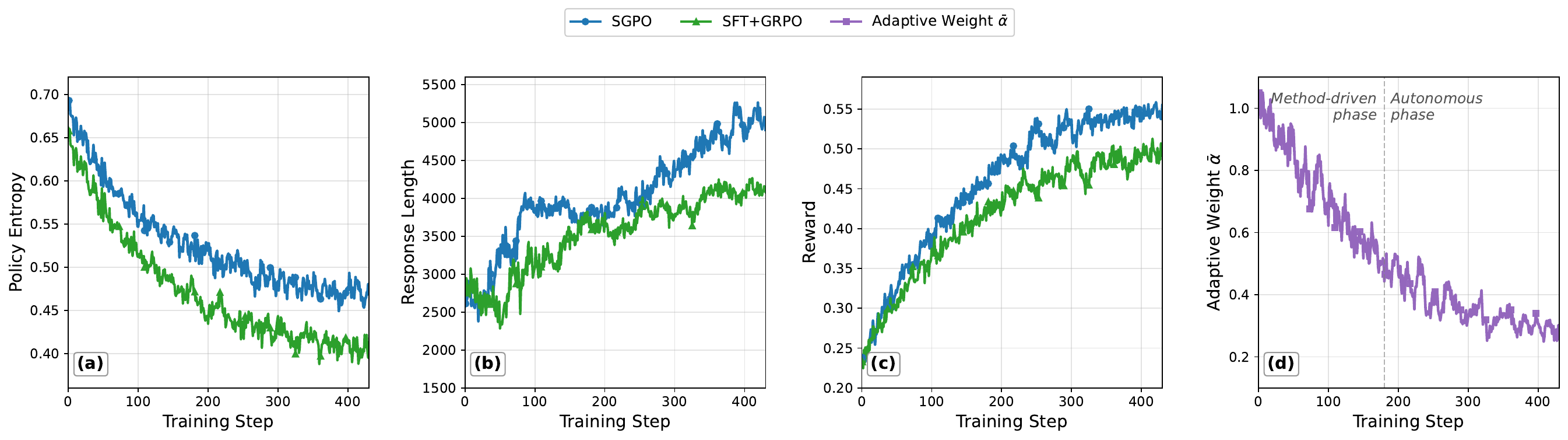}
    \caption{Training dynamics of SGPO and SFT+GRPO.
    (a)~Policy entropy: SGPO maintains consistently higher entropy.
    (b)~Response length: SGPO reaches longer, more structured
        responses faster.
    (c)~Reward: SGPO achieves a higher final reward.
    (d)~Average adaptive weight $\bar{\alpha}$: the weight decreases
        as the autonomous policy improves, reflecting a natural
        transition from strategy-driven exploration to autonomous
        optimization.}
    \label{fig:training_dynamics}
    \vspace{-10pt}
\end{figure*}

\subsection{Further Analysis}

\subsubsection{Training Dynamics}

Figure~\ref{fig:training_dynamics}(a)(b)(c) compares SGPO with
SFT+GRPO. SGPO maintains higher policy entropy throughout training.
Two factors likely contribute: the KL distillation objective introduces
new high-probability options at strategy-critical positions,
enriching the unguided policy's distribution, while the proximal
constraints prevent this distillation from over-concentrating the
distribution. Both response length and reward increase rapidly in early
training for both methods, but SGPO exhibits faster growth, suggesting
that strategy conditioning helps the policy discover structured
solution patterns earlier.

Figure~\ref{fig:training_dynamics}(d) tracks the average adaptive weight
$\bar{\alpha}$ across training steps. In early training, the gap
between guided and autonomous pass rates is large, yielding high
$\bar{\alpha}$ and strong distillation pressure. As the autonomous
policy improves and the pass-rate gap narrows, $\bar{\alpha}$ decreases
steadily. This confirms the intended training dynamics: the framework
transitions naturally from \emph{strategy-driven fast improvement} in
early stages to \emph{autonomous-policy-dominated steady optimization}
in later stages, with no manual scheduling required.

\begin{figure*}[t]
    \centering
    \includegraphics[width=\textwidth]{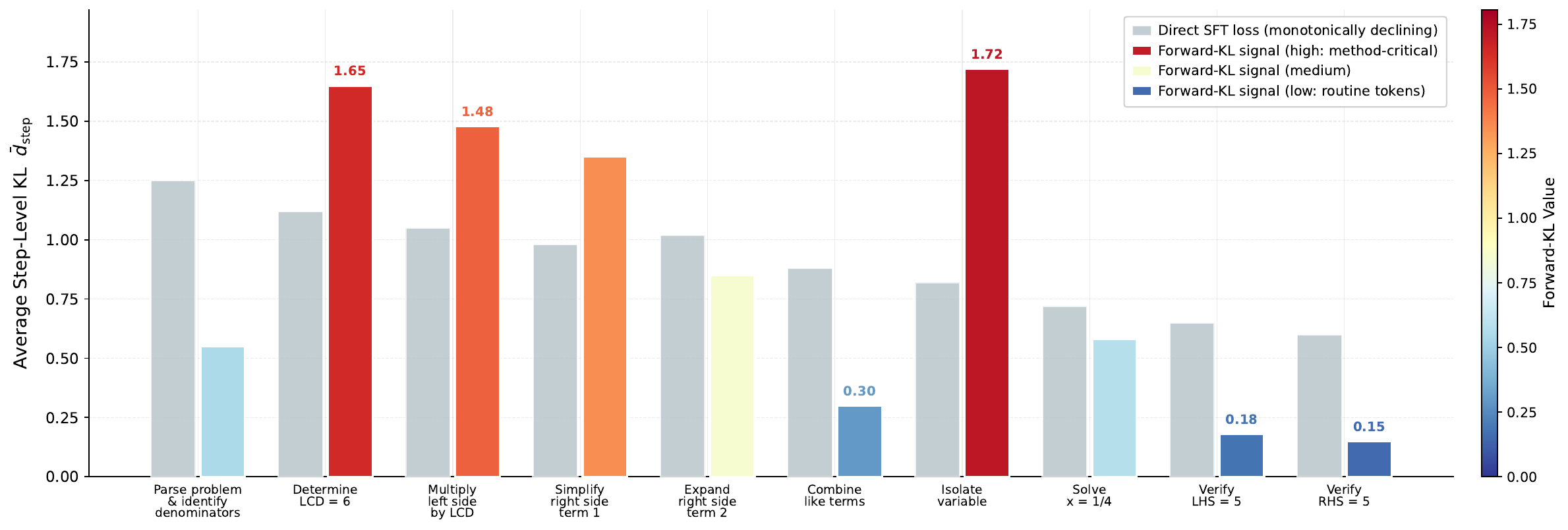}
    \caption{Step-level KL analysis on a training example. Gray bars:
    direct SFT loss, declining monotonically. Colored bars: forward-KL
    signal, peaking at strategy-critical steps (red, e.g., determining
    the LCD, isolating the variable) and minimal at routine steps
    (blue).}
    \label{fig:reasoning_step_kl}
    \vspace{-15pt}
\end{figure*}

\begin{table}[htbp]
\centering
\caption{KL-based distillation vs.\ direct SFT on successful
strategy-guided trajectories.}
\label{tab:dense_reward_analysis}
\small
\setlength{\tabcolsep}{6pt}
\resizebox{0.75\linewidth}{!}{
\begin{tabular}{lccccc}
\toprule
\textbf{Distillation type} & \textbf{MATH500} & \textbf{AMC23} & \textbf{Olympiad} & \textbf{AIME24} & \textbf{Average} \\
\midrule
KL distillation & \textbf{82.7} & \textbf{55.9} & \textbf{50.0} & \textbf{19.7} & \textbf{52.1} \\
Direct SFT & 79.8 & 53.0 & 47.0 & 17.3 & 49.3 \\
\bottomrule
\end{tabular}
}
\vspace{-10pt}
\end{table}

\subsubsection{Selective Amplification of Strategy-Critical Tokens}
\label{sec:selective}

\begin{wrapfigure}{r}{0.4\textwidth}
    \vspace{-15pt}
    \centering
    \includegraphics[width=0.4\textwidth]{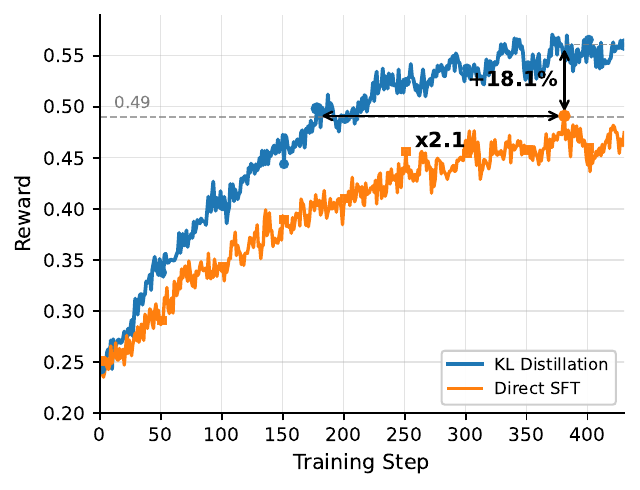}
    \caption{KL-based distillation yields faster convergence and a
    higher reward ceiling than direct SFT on guided trajectories.}
    \label{fig:dense_reward_effectiveness}
    \vspace{-15pt}
\end{wrapfigure}

To validate that KL-based distillation is more effective than uniform
imitation, we replace the forward-KL objective with direct SFT on the
same successful strategy-guided trajectories, keeping all other components
identical. Table~\ref{tab:dense_reward_analysis} shows that KL-based
distillation outperforms direct SFT by 2.8 points on average (52.1 vs.\
49.3), and Figure~\ref{fig:dense_reward_effectiveness} confirms that
the advantage holds throughout training in both convergence speed and
final ceiling.

The performance gap arises from the granularity of the learning signal.
Direct SFT applies uniform fitting pressure across all tokens, whereas
the KL objective automatically concentrates optimization on tokens whose
generation probability shifts most under strategy conditioning.
Figure~\ref{fig:reasoning_step_kl} illustrates this on a training
example involving a linear equation with fractional coefficients, where
none of the 8 autonomous samples are correct while 2 of 4 strategy-guided
samples succeed. We segment the successful trajectory into ten semantic
steps and compute the average KL within each step. The KL
signal peaks at strategy-critical steps such as determining the least
common denominator and distributing it across both sides, all
corresponding directly to the extracted strategy. In contrast, the
SFT loss declines monotonically across steps, reflecting generic token
prediction difficulty rather than strategy relevance. This confirms
that the forward-KL objective provides an inherently strategy-aware signal
that distinguishes strategic decisions from routine verbalization.

\vspace{-5pt}
\section{Conclusion}
\vspace{-5pt}
We presented SGPO, a framework that distills reusable strategy
descriptions from strong-model responses and transfers them into the
target model's reasoning policy through token-level forward-KL
distillation between strategy-guided and unguided distributions. Combined
with autonomous GRPO exploration, proximal constraints at the trajectory
and token levels, and adaptive instance-level weighting, the framework
consistently outperforms strong baselines across mathematical
benchmarks and model families.

Our analysis further shows that the forward-KL objective provides an
inherently selective distillation signal that concentrates on
strategy-critical tokens without requiring any explicit annotation,
offering a more effective alternative to uniform trajectory imitation.
Moreover, strategy distillation and base model capability scale
complementarily, with gains increasing as the model's foundational
reasoning competence grows. Together, these findings suggest that
transferring knowledge at the strategy level, rather than the
trajectory level, represents a promising direction for reasoning
distillation.

\section*{Acknowledgments}
This work was supported by the National Natural Science Foundation of China under Grant No.~62576368.

\section*{Ethics Statement}
This work focuses on improving the reasoning capabilities of language
models through strategy distillation. All training data are drawn
from publicly available mathematical problem datasets, and no human
subjects or personally identifiable information are involved. The
strong-model responses used for strategy extraction are generated by
existing publicly accessible models. We acknowledge that improved
reasoning capabilities could in principle be misused, but the
mathematical reasoning setting studied here poses minimal direct risk.
We will release our code and strategy extraction prompts to support
reproducibility.

\bibliography{colm2026_conference}
\bibliographystyle{colm2026_conference}

\appendix

\section{Limitations}
Several limitations point to future directions. First, our experiments
focus on mathematical reasoning; validating the framework on other
reasoning domains such as code generation, logical reasoning, and
scientific problem solving remains important. Second, strategy
extraction currently relies on a strong model and is not fully quality-controllable; developing more robust extraction methods or enabling
the student model to discover strategies autonomously would reduce
this dependence. Third, the complementary scaling pattern implies that very weak base models may not benefit substantially from strategy distillation, and understanding the precise capability threshold deserves further investigation. Finally, scaling the framework to larger training sets and stronger base models would help clarify its broader applicability.


\section{Strategy Extraction Pipeline}
\label{app:strategy_extraction}

Strategy descriptions are extracted as a preprocessing step before
RL training. Given a strong-model (DeepSeek-R1) response to each
training problem, we prompt the same model to produce a strategy
description conforming to the three-component structure described in
\S\ref{sec:formulation}: problem type identification, strategy
selection, and general procedural steps. To mitigate extraction
variance, we sample five candidate descriptions per response and rank
them with a separate scoring prompt that evaluates four criteria:
(i)~correctness of problem type identification, (ii)~appropriateness
and specificity of the chosen strategy, (iii)~abstraction level of the
procedural steps (neither too vague nor too detailed), and (iv)~absence
of intermediate computations or the final answer. The highest-scoring
candidate is retained. Prompt templates are given in
Appendix~\ref{app:prompts}.

\begin{table*}[h]
\centering
\caption{Prompt template for extracting strategy descriptions.}
\label{tab:method_extract_prompt}
\small
\renewcommand{\arraystretch}{1.1}
\resizebox{\textwidth}{!}{
\begin{tabular}{l}
\toprule
\begin{tabular}[c]{@{}l@{}}
You are given a math problem together with a complete solution response written by a strong reasoning model.\\
Your task is to extract the problem-solving \textbf{strategy} behind the response. The extracted strategy\\
must contain exactly three components:\\
1. \textbf{Problem type}: classify the problem into a specific, recognizable category.\\
2. \textbf{Strategy}: state the specific principle, theorem, or technique used to solve this type of problem.\\
3. \textbf{Procedural steps}: list the high-level steps for executing the strategy, in order.\\
\\
Do \textbf{not} include intermediate numerical computations or the final answer.\\
The output should be concise, generalizable, and helpful for guiding another model to solve similar problems.\\
\\
Output format:\\
Problem type: [...]\\
Strategy: [...]\\
Steps: (a) ... (b) ... (c) ...
\end{tabular} \\
\midrule
\textbf{Problem:} [problem text] \\
\textbf{Response:} [strong-model solution] \\
\midrule
\textbf{Strategy:} \\
\bottomrule
\end{tabular}
}
\end{table*}

\section{Prompt Templates}
\label{app:prompts}

\begin{table*}[h]
\centering
\caption{Prompt template for scoring candidate strategy descriptions.}
\label{tab:method_score_prompt}
\small
\renewcommand{\arraystretch}{1.1}
\resizebox{\textwidth}{!}{
\begin{tabular}{l}
\toprule
\begin{tabular}[c]{@{}l@{}}
You are given a math problem, a reference solution, and several candidate extracted strategies.\\
Assign a quality score (0.0--1.0) to each candidate based on the following criteria:\\
1. Whether the problem type is correctly and specifically identified.\\
2. Whether the chosen strategy is appropriate, specific, and non-trivial.\\
3. Whether the procedural steps are at the right abstraction level (not too vague, not too detailed).\\
4. Whether the strategy avoids revealing intermediate computations or the final answer.\\
\\
Output format: Score: [score1, score2, ...]
\end{tabular} \\
\midrule
\textbf{Problem / Response / Candidate Strategies} (omitted for brevity) \\
\midrule
\textbf{Score:} \\
\bottomrule
\end{tabular}
}
\end{table*}

\section{Training Algorithm}
\label{app:algorithm}

Algorithm~\ref{alg:hybrid_method_distill} summarizes the full
procedure. For each problem, we jointly sample an autonomous group and
a strategy-guided group. The autonomous group is optimized through GRPO.
The guided group contributes only when it contains correct trajectories,
in which case the most reachable correct response (scored under the
\emph{unguided} policy) is selected for proximal KL distillation with
adaptive weighting.

\begin{algorithm}[htbp]
\caption{Strategy-Guided Policy Optimization (SGPO)}
\label{alg:hybrid_method_distill}
\begin{algorithmic}[1]
\Require Problem distribution $\mathcal{P}(Q)$, policy $\pi_\theta$,
         strategy dataset $\mathcal{D}_m=\{(q,s)\}$,
         group sizes $G_1,G_2$
\Ensure  Updated policy $\pi_\theta$
\For{each training iteration}
    \State Sample minibatch $\{q^{(b)}\}_{b=1}^{B}\sim\mathcal{P}(Q)$
    \For{each problem $q^{(b)}$}
        \State Retrieve strategy $s^{(b)}$
        \State Sample autonomous group
               $\{o_i\}_{i=1}^{G_1}\sim\pi_\theta(\cdot\mid q^{(b)})$
        \State Sample strategy-guided group
               $\{\tilde{o}_j\}_{j=1}^{G_2}\sim
                \pi_\theta(\cdot\mid q^{(b)},s^{(b)})$
        \State Compute rewards $\{r(o_i,q^{(b)})\}$ and
               $\{r(\tilde{o}_j,q^{(b)})\}$
        \State Compute GRPO advantages $\{\hat{A}_i\}$ and loss
               $\mathcal{L}_{\mathrm{GRPO}}$
        \State $\mathcal{C}\leftarrow\{j: r(\tilde{o}_j,q^{(b)})=1\}$
        \If{$\mathcal{C}\neq\emptyset$}
            \State Select reachable target:
                   $o_m^{+}=\arg\max_{\tilde{o}_j\in\mathcal{C}}
                    \sum_t\log\pi_\theta(\tilde{o}_{j,t}\mid
                    q^{(b)},\tilde{o}_{j,<t})$
                   \Comment{unguided log-prob}
            \State Compute clipped KL loss
                   $\mathcal{L}_{\mathrm{KD}}^{\mathrm{clip}}$
                   on $o_m^{+}$
            \State Compute adaptive weight $\alpha(q^{(b)})$
                   via Eq.~\ref{eq:adaptive_weight}
        \Else
            \State $\mathcal{L}_{\mathrm{KD}}^{\mathrm{clip}}
                    \leftarrow 0$,\;
                   $\alpha(q^{(b)})\leftarrow 0$
        \EndIf
        \State $\mathcal{L}^{(b)}\leftarrow
               \mathcal{L}_{\mathrm{GRPO}}
               +\alpha(q^{(b)})\,
                \mathcal{L}_{\mathrm{KD}}^{\mathrm{clip}}$
    \EndFor
    \State Update $\theta$ by gradient descent on
           $\frac{1}{B}\sum_{b}\mathcal{L}^{(b)}$
\EndFor
\end{algorithmic}
\end{algorithm}

\section{Implementation Details}
\label{app:impl_details}

\textbf{Hardware and frameworks.}
All experiments are conducted on 4 nodes, each equipped with 8 GPUs
(32 GPUs in total). We use the veRL framework~\citep{verl} for all
training (both SFT and RL stages) and vLLM~\citep{vllm} for inference
during evaluation. Answer correctness is verified using
Math-Verify~\citep{Math-Verify}. End-to-end training (SFT warm-up +
RL) takes approximately 17 hours for the 7B model.

\textbf{SFT warm-up.}
We use roughly one third of the 8.5K training instances for supervised
warm-up to endow the base models with basic thinking ability. We use batch size 64, train for 10 epochs with learning rate
$1{\times}10^{-5}$, warm-up ratio 0.1, and maximum sequence length
16K. The checkpoint with the lowest loss on a randomly sampled
one-tenth validation split is used to initialize RL.

\textbf{Reinforcement learning.}
The total group size is 12 ($G_1{=}8$ autonomous, $G_2{=}4$
strategy-guided). We use temperature 1.0, top-$p{=}0.95$, maximum
generation length 16K, and train for 10 epochs with learning rate
$1{\times}10^{-6}$. Hyperparameters: GRPO clipping bounds
$\varepsilon_{\text{low}}{=}0.2$ and $\varepsilon_{\text{high}}{=}0.28$~\citep{dapo},
KL threshold $\delta{=}1.0$, base adaptive coefficient
$\alpha_0{=}0.5$, smoothing constant $\gamma{=}0.1$, and maximum
adaptive weight $\alpha_{\max}{=}0.8$.

\textbf{Baselines.}
SFT and SFT+GRPO use the same warm-up hyperparameters as our method.
For HPT and LUFFY, we match the total number of samples and sampling
parameters to ensure fair comparison, while keeping their
method-specific hyperparameters at the default values reported in the
original papers, which already yield stable performance in our setting.

\section{KL Direction Ablation}
\label{app:kl_direction_ablation}

We compare the forward KL direction
$D_{\mathrm{KL}}(\pi_{\text{guided}}\,\|\,\pi_{\text{unguided}})$
used in our method with the reverse direction
$D_{\mathrm{KL}}(\pi_{\text{unguided}}\,\|\,\pi_{\text{guided}})$
on Qwen2.5-7B-Instruct.

\begin{table}[htbp]
\centering
\caption{KL direction ablation on Qwen2.5-7B-Instruct.}
\label{tab:kl_direction}
\small
\setlength{\tabcolsep}{6pt}
\resizebox{0.7\linewidth}{!}{
\begin{tabular}{lccccc}
\toprule
\textbf{KL direction} & \textbf{MATH500} & \textbf{AMC23} & \textbf{Olympiad} & \textbf{AIME24} & \textbf{Average} \\
\midrule
Forward (ours) & \textbf{82.7} & \textbf{55.9} & \textbf{50.0} & 19.7 & \textbf{52.1} \\
Reverse & 81.3 & 54.2 & 48.2 & \textbf{20.0} & 50.9 \\
\bottomrule
\end{tabular}
}
\end{table}

The forward KL consistently outperforms the reverse direction. The
strategy-guided distribution reflects a particular strategy, but a problem
may admit multiple effective strategies that the model has already
discovered through autonomous exploration. Forward KL only requires
coverage of the guided distribution, preserving these existing
alternatives while absorbing the new methodological insight. Reverse KL
would instead collapse the unguided policy onto the guided modes,
risking loss of alternative strategies.

\section{Adaptive Weighting Strategy Comparison}
\label{app:weight_ablation}

We compare the pass-rate-based adaptive weighting
(Eq.~\ref{eq:adaptive_weight}) against two alternatives on
Qwen2.5-7B-Instruct: (i)~a fixed coefficient $\alpha{=}1.0$ (equivalent
to the ``w/o adaptive weighting'' setting in Table~\ref{tab:ablation}),
and (ii)~a group-advantage weight described below.

\textbf{Group-advantage weight.}\quad
Let $o^+$ be the selected correct strategy-guided trajectory and
$\{o_i\}_{i=1}^{G_1}$ the autonomous group. We merge them into a
combined group $\mathcal{G}=\{o_1,\dots,o_{G_1},o^+\}$ and compute a
GRPO-style normalized advantage for the guided trajectory:
\begin{equation}
\hat{A}_{m^+}
= \frac{R(o^+,q)-\mu_{\mathcal{G}}}
       {\sigma_{\mathcal{G}}+\epsilon},
\label{eq:group_adv}
\end{equation}
where $\mu_{\mathcal{G}}$ and $\sigma_{\mathcal{G}}$ are the mean and
standard deviation of rewards within $\mathcal{G}$. The distillation
weight is then set to
$\alpha_{\text{group}}(q)=\mathrm{clip}(\hat{A}_{m^+},\,0,\,
\alpha_{\max})$.

\begin{table}[htbp]
\centering
\caption{Comparison of weighting strategies on Qwen2.5-7B-Instruct.}
\label{tab:weight_comparison}
\small
\setlength{\tabcolsep}{6pt}
\resizebox{0.75\linewidth}{!}{
\begin{tabular}{lccccc}
\toprule
\textbf{Weighting strategy} & \textbf{MATH500} & \textbf{AMC23} & \textbf{Olympiad} & \textbf{AIME24} & \textbf{Average} \\
\midrule
Pass-rate-based (ours) & \textbf{82.7} & \textbf{55.9} & \textbf{50.0} & \textbf{19.7} & \textbf{52.1} \\
Fixed $\alpha{=}1.0$ & 81.0 & 54.8 & 48.9 & 18.7 & 50.9 \\
Group-advantage & 81.4 & 54.5 & 48.4 & 18.0 & 50.6 \\
\bottomrule
\end{tabular}
}
\end{table}

The pass-rate-based design consistently outperforms both alternatives.
The fixed coefficient applies uniform distillation strength regardless
of instance difficulty. The group-advantage weight reflects the relative
quality of the guided trajectory within the combined group but conflates
two distinct signals: problem difficulty and strategy effectiveness.
A correct guided trajectory receives a high advantage whenever the
autonomous group mostly fails, yet this does not indicate how much of
the success is attributable to the strategy itself versus the
inherent difficulty of the problem. In contrast, the pass-rate-based
design disentangles these factors by directly measuring the pass-rate
gap between guided and autonomous groups, providing a more faithful
estimate of the marginal benefit of strategy guidance.

\section{Strategy Description Quality Analysis}
\label{app:method_quality}

We randomly sample 500 strategy descriptions and score each using
Step-3.5-Flash~\citep{step35flash} on the same four criteria used in the extraction scoring prompt
(Appendix~\ref{app:prompts}): correctness of problem type
identification, appropriateness of strategy, abstraction level of
procedural steps, and absence of answer leakage. Each criterion is
rated on a 0--1 scale and the four scores are averaged into an overall
quality score. Table~\ref{tab:method_quality_dist} reports the
distribution.

\begin{table}[htbp]
\centering
\caption{Quality distribution of extracted strategy descriptions.}
\label{tab:method_quality_dist}
\small
\setlength{\tabcolsep}{8pt}
\begin{tabular}{lccc}
\toprule
\textbf{Quality band} & \textbf{Score range} & \textbf{Count} & \textbf{\%} \\
\midrule
High   & $[0.8, 1.0]$ & 197 & 39.4 \\
Medium & $[0.5, 0.8)$ & 272 & 54.4 \\
Low    & $[0.0, 0.5)$ & 31 & 6.2 \\
\bottomrule
\end{tabular}
\end{table}

The majority of extracted descriptions fall in the High and Medium
bands, indicating that the extraction pipeline combined with the
scoring-based selection (Appendix~\ref{app:strategy_extraction})
produces generally reliable strategy descriptions. The small fraction
of Low-quality descriptions is handled by the adaptive weighting
mechanism (Eq.~\ref{eq:adaptive_weight}): when a strategy is
unhelpful, the guided pass rate $p_{\text{guide}}$ will not
substantially exceed $p_{\text{auto}}$, automatically reducing
$\alpha(q)$ and limiting the distillation signal from these instances.

\end{document}